\documentclass{svproc}
\usepackage{graphicx}%
\usepackage{multirow}%
\usepackage{amsmath,amssymb,amsfonts}%
\usepackage{mathrsfs}%
\usepackage[title]{appendix}%
\usepackage{xcolor}%
\usepackage{textcomp}%
\usepackage{manyfoot}%
\usepackage{booktabs}%
\usepackage{algorithm}%
\usepackage{algorithmicx}%
\usepackage{algpseudocode}%
\usepackage{listings}%
\usepackage{url}

\begin{document}
\title{DCOR: Anomaly Detection in Attributed Networks via Dual Contrastive Learning Reconstruction }
\titlerunning{Dual Contrastive Learning Reconstruction}
\author{
Hossein Rafieizadeh\textsuperscript{1}, 
Hadi Zare\textsuperscript{2}, 
Mohsen Ghassemi Parsa\textsuperscript{3}, 
Hadi Davardoust\textsuperscript{4}, 
Meshkat Shariat Bagheri\textsuperscript{5}
}

\authorrunning{H. Rafieizadeh et al.}

\institute{
Department of Data Science and Technology, School of Intelligent Systems Engineering, University of Tehran \\[0.3cm]
\textsuperscript{1}\email{hossein.rafiee@ut.ac.ir}, 
\textsuperscript{2}\email{h.zare@ut.ac.ir}, 
\textsuperscript{3}\email{mgparsa@ut.ac.ir}, 
\textsuperscript{4}\email{davardoust@ut.ac.ir}, 
\textsuperscript{5}\email{m.shariatbagheri@ut.ac.ir}
}

\maketitle           

\vspace{-0.6cm} 

\begin{abstract}

Anomaly detection using a network-based approach is one of the most efficient ways to identify abnormal events such as fraud, security breaches, and system faults in a variety of applied domains. While most of the earlier works address the complex nature of graph-structured data and predefined anomalies, the impact of data attributes and emerging anomalies are often neglected. This paper introduces DCOR, a novel approach on attributed networks that integrates reconstruction-based anomaly detection with Contrastive Learning. Utilizing a Graph Neural Network (GNN) framework, DCOR contrasts the reconstructed adjacency and feature matrices from both the original and augmented graphs to detect subtle anomalies. We employed comprehensive experimental studies on benchmark datasets through standard evaluation measures. The results show that DCOR significantly outperforms state-of-the-art methods. Obtained results demonstrate the efficacy of proposed approach in attributed networks with the potential of uncovering new patterns of anomalies.

\end{abstract}

\section{Introduction}

Anomaly detection in attributed networks plays a critical role in modern networked systems due to their increasing complexity and interconnectivity \cite{1}.
Anomalies often occur in applied domains such as financial systems, computer networks, and the internet known as fraud, security breaches, or system faults with billions of costs for companies \cite{2}.

Traditional anomaly detection methods often have problems when facing with attributed graph-structured data. Techniques such as Local Outlier Factor (LOF) \cite{3} and Structural Clustering Algorithm for Networks (SCAN) \cite{4} primarily rely on local density and structural deviations.  
However, these approaches suffered to deal with node attributes and connectivity structure of the network simultaneously.

The rise of deep learning brought new technologies in the identification of anomalies, such as autoencoders and Graph Neural Networks (GNNs). AnomalyDAE \cite{5} uses dual autoencoders to reconstruct both the adjacency matrix and node features for detecting anomalies that hurt either the structure or network attributes. Recent works still build upon those foundations, like Dominant \cite{6}, which demonstrated good results on attributed networks based on graph convolutional neural networks.

Additionally, contrastive learning is a robustness-enhancing method, which learns to distinguish between augmented versions of data. It has shown promise within graph representation learning, for example, GraphCL \cite{7}, which uses data augmentation in order to create contrasting views of graphs. That is, contrastive learning maximizes the agreement among different augmented views of the same data (positive pairs) while minimizing the agreement between the views of different data points (negative pairs). The aforementioned pre-training strategies for GNNs were experimented upon by Hu et al. \cite{8}, who showed the capability of the model to capture global and local structural information.

Based on these insights, we propose a new way of integrating dual reconstruction-based anomaly detection with contrastive learning. Our method, called Dual Reconstruction Contrastive Learning (DCOR), introduces a GNN-based framework that contrasts the reconstructed adjacency and feature from both original and augmented graphs. This contrast allows DCOR to better adapt in capturing both structural and attribute anomalies, due to the learned robust reconstructions through contrastive learning, ensuring that even subtle anomalies missed by previous methods are detected. Furthermore, this contrast enhances the quality of our reconstruction, which in turn improves the overall performance of the model. By increasing the quality of the main network reconstruction, we indirectly enhance the representations and consequently strengthen the model's ability to detect anomalies.

In this paper, we present the design and implementation of DCOR, along with a detailed performance evaluation across multiple datasets. After summarizing recent related work, we define our proposed approach, which includes the graph augmentation process, the anomaly detection model, and the contrastive learning framework.

We evaluate DCOR's performance against several baseline methods using five datasets from diverse domains, including Flickr, Amazon, Enron, and Facebook. The results demonstrate that DCOR achieves notable improvements compared to state-of-the-art methods.

\section{Problem Definition}

Formally, let \( G = (A, X) \) represent an attributed network, where \( A \in \mathbb{R}^{n \times n} \) is the adjacency matrix and \( X \in \mathbb{R}^{n \times d} \) is the attribute matrix. Each node \( v_i \) in the graph is represented by a row in both \( A \) and \( X \). The adjacency matrix \( A \) captures the structural relationships between nodes, with \( A_{ij} = 1 \) if there is an edge between nodes \( v_i \) and \( v_j \), and \( A_{ij} = 0 \) otherwise. The attribute matrix \( X \) contains attribute vectors for each node, where \( d \) is the dimensionality of the attribute space.

The goal of anomaly detection is to recognize a subset of nodes \( V_{\text{anomaly}} \subseteq V \) where each node \( v \in V_{\text{anomaly}} \) shows anomalous behavior. This anomalous behavior could happen in three ways: structural anomalies, attribute anomalies, and jointly together.

Specifically, we define our anomaly detection problem as follows:

\begin{itemize}
    \item \textbf{Input}: An attributed network \( G = (A, X) \) with adjacency matrix \( A \) and node attribute matrix \( X \).
    \item \textbf{Output}: A ranked ordering of nodes via their anomaly scores, with higher scores for a more likely of being anomalous.
\end{itemize}

Our proposed method aims to produce an anomaly score \( S(v) \) for each node \( v \in V \), where nodes with the highest scores are identified as anomalies. 

\section{Related Work}

\subsection{Autoencoder and GNN-Based Anomaly Detection}  

Some works combine autoencoders with GNNs for better anomaly detection because they capture complex dependencies in graph data. For instance, Zhou and Liu proposed AnomalyDAE \cite{5}, which is a dual autoencoder architecture to simultaneously reconstruct the adjacency matrix and node features for anomaly detection. Some analogical approaches, such as Dominant \cite{6}, based on GNNs, try to learn node embeddings from a GNN to acquire good performance in node-wise anomaly detection. A recent and likely the most relevant one is by Liu et al. \cite{9}, presenting contrastive learning on attributed graphs to improve anomaly detection. The focus of such an approach lies in generating effective negative samples to distinguish normal patterns from anomalous ones, thereby achieving enhanced robustness of the model. Such approaches underline the strength of autoencoders and GNNs in capturing structural nuances and dependencies in graph-based tasks of anomaly detection.

\subsection{Graph Contrastive Learning and Self-Supervised Learning}

Some of the latest techniques in this line, which have considerably advanced the robustness and generalization of GNNs, are graph contrastive learning and self-supervised learning with data augmentation.

\textbf{Graph Contrastive Learning:} GraphCL \cite{7} introduces a way to improve the GCNs by providing different views of graphs with respect to data augmentations, like node dropping and edge perturbations. It works in a way in which the augmented views between the same graph are brought to maximum agreement, but minimized with those of different graphs.

\textbf{Self-Supervised Learning with Data Augmentation:} In \cite{10}, robust embeddings of anomaly detection are generated by adding node and graph augmentations. Curriculum negative sampling is introduced to adjust the difficulty of sampling, which has resulted in strengthening the model's ability to improve the identification of anomalous patterns \cite{11}.

Some other recent works: adversarial training of GCL to enhance robustness in \cite{12}; a self-supervised framework for integrating graph augmentation and multi-view learning in \cite{13}; an SSL-over-graph federated learning approach in \cite{14}. Some strategies for pre-training GNNs are also discussed in reference \cite{8} for node and graph-level tasks.

\subsection{Recent Advances in Graph Anomaly Detection}

A dynamic graph anomaly detection using GNNs was proposed in \cite{15}, followed by the introduction of the first contrastive learning method for detecting anomalies in temporal networks. A domain-specific contrastive learning technique that is agnostic to the graph was proposed in \cite{16}. They proposed a hybrid model that combined graph neural networks with reinforcement learning for anomaly detection within dynamic graphs.

\section{Proposed Method}

Our method starts by generating both the original network and an augmented version. We then focus on comparing the reconstructed adjacency and feature matrices from the original network with those from the augmented one.

What sets our approach apart from traditional contrastive learning frameworks \cite{17}, which usually compare embeddings from original and augmented networks, is our focus on reconstructing the actual graph data. By shifting attention to reconstruction, we can uncover more intricate anomalies and subtle differences in the graph's structure and features—details that might be missed if we were only looking at the embeddings. Our main goal is to refine the quality of these reconstructions within the autoencoder, allowing the model to more effectively distinguish between normal and anomalous data by emphasizing the key differences.

Initially, we use an autoencoder to help us in the reconstruction of the adjacency and feature matrices of the graph. These reconstructions are the basis for carrying out anomaly detection. Now begins the process of contrastive learning. At this stage, the model is going to compare the reconstructed adjacency and feature matrices of the original and augmented graphs. The aim is to augment the accuracy of anomaly detection through this contrast. Such an approach enables the model to detect slight differences in the structure and features of normal versus anomalous graphs, thus making anomaly detection much more accurate.

During reconstruction, anomalies found in the augmented graph lead the model to try deviating from their reconstruction of the original network. At the same time, the model will also try keeping its reconstruction for the nodes that are anomalously far away from theirs. For instance, if an augmented network node is infected, it tries to predict the reconstruction of that node back toward its corresponding node in the original network. Such contrasts are made both in terms of node features and the adjacency matrix. Such contrast will therefore let the model learn how it should accordingly reconstruct the graph. It implies that the model would be more skilled at detecting anomalies because differences and similarities at both the feature and structural level between normal and abnormal nodes could be distinguished. Ultimately, we are interested in improving the power of the model to capture subtle yet significant anomalies present within graph data by ameliorating the quality of the reconstructions supplied by the autoencoder.

\subsection{Graph Data Augmentation}

Anomaly detection graph augmentation refers to the incorporation of artificially generated anomalies into graph data to enhance the effectiveness of detection algorithms \cite{9}. In order to enable effective testing and verification, we incorporate unconventional nodes or edges into the network, hence producing concrete abnormalities. Both structural and feature anomalies are included, with augmentation rates optimally adjusted to the unique characteristics of each dataset. Upon encountering an anomaly, a node is allocated the label of anomalous (anomaly label), which facilitates straightforward discrimination during testing. By incorporating tailored anomaly creation, the model's robustness and capacity to generalize to real-world scenarios are improved.

Graph augmentation can be expressed as:
\begin{equation}
G' = G + A
\label{eq:augmentation}
\end{equation}
where $G'$ is the augmented graph, $G$ is the original graph, and $A$ the anomalies introduced into the graph.

\subsubsection{Node-Level Augmentation}
We perturb node features by adding noise to simulate anomalies and increase data diversity. Perturbation improves the model's ability to generalize. Techniques include:

\begin{itemize}

\item \textbf{Feature Copying:} Replicating the feature vector of a node with maximum deviation to a target node creates attribute anomalies \cite{18}, useful in detecting subtle deviations.

\item \textbf{Feature Scaling:} Adjusting feature values by a scaling factor creates anomalies in feature distribution, enhancing detection.

\end{itemize}

\subsubsection{Structural Augmentation}
Structural augmentation alters the graph by adding or removing edges to simulate unusual connectivity patterns. This helps the model generalize and detect structural anomalies. Techniques include:

\begin{itemize}

\item \textbf{Adding Subgraph:} Introducing dense subgraphs around nodes creates structural anomalies, simulating unusual connectivity patterns.

\item \textbf{Drop all connections:} Disconnecting a node isolates it, simulating scenarios like faults or network issues, and test the model's ability to detect isolated nodes.

\end{itemize}

Our approach customizes augmentation rates based on each dataset's characteristics, rather than using uniform strategies. This improves anomaly detection performance by optimizing model robustness and generalization.

\subsection{Anomaly Detection Model and Contrastive Learning}

Traditional contrastive learning frameworks \cite{17} focus on comparing embeddings from different augmented views of the same data to maximize agreement between similar pairs and minimize agreement between dissimilar pairs. Our approach builds upon this concept by first reconstructing the graph data using a dual autoencoder model. Since typical representations may lose some sensitive information about the nodes, which can hinder accurate contrasts, we compare the reconstructed adjacency matrices and feature matrices directly. By feeding both the main graph and the augmented graph into the dual autoencoder, we enhance the detection of anomalies and ensure more precise results.

\subsubsection{Dual Autoencoder Model}

The dual autoencoder model reconstructs the graph's adjacency and feature matrices using both structure and attribute autoencoders. The main and augmented graphs are fed into the model to obtain reconstructed matrices.

\paragraph{Structure Autoencoder}

The structure autoencoder encodes the graph structure into a latent representation and reconstructs the adjacency matrix.

- \textbf{Encoder}: Compresses the input graph \( (A, X) \) into a latent representation \( Z_V \):
  \begin{equation}
  \tilde{Z}_V = \sigma(XW_V^{(1)} + b_V^{(1)})
  \end{equation}
where \( X \) is the feature matrix, \( W_V^{(1)} \) and \( b_V^{(1)} \) are weight and bias parameters, and \( \sigma \) is the activation function.

The attention scores (\(e_{ij}\)) between nodes \(i\) and \(j\) are computed as follows:
\begin{equation}
e_{ij} = \text{attn}(\tilde{Z}_i^V, \tilde{Z}_j^V) = \sigma\left( a^T \cdot [W_V^{(2)} \tilde{Z}_i^V \,||\, W_V^{(2)} \tilde{Z}_j^V] \right)
\end{equation}
where \( \tilde{Z}_i^V \) and \( \tilde{Z}_j^V \) are the latent representations of nodes \(i\) and \(j\), \(W_V^{(2)}\) is the weight matrix, \(a\) is the attention weight vector, and \([\,||\,]\) represents concatenation of the vectors.

A graph attention layer then aggregates the representation from neighboring nodes:
\begin{equation}
\gamma_{ij} = \frac{\exp(e_{ij})}{\sum_{k \in N_i} \exp(e_{ik})}
\end{equation}
where \( \gamma_{ij} \) is the attention coefficient for nodes \(i\) and \(j\), computed based on the attention scores \(e_{ij}\).

The final embedding for node \(i\), denoted as \( Z_V^i \), is computed as:
\begin{equation}
Z_V^i = \sum_{k \in N_i} \gamma_{ik} \tilde{Z}_V^k
\end{equation}
where \( N_i \) represents the neighbors of node \(i\).

- \textbf{Decoder}: Reconstructs the adjacency matrix \( \hat{A} \) from \( Z_V \):
  \begin{equation}
  \hat{A} = \text{Sigmoid}(Z_V Z_V^T)
  \end{equation}
  where \( Z_V \) is the latent node representation and \( \hat{A} \) is the reconstructed adjacency matrix.

\paragraph{Attribute Autoencoder}

The attribute autoencoder encodes node attributes into latent space and reconstructs the attribute matrix.

- \textbf{Encoder}: Transforms node attributes \( X \) into a latent representation \( Z_A \):
  \begin{equation}
  \tilde{Z}_A = \sigma(X^T W_A^{(1)} + b_A^{(1)})
  \end{equation}
  where \( W_A^{(1)} \) and \( b_A^{(1)} \) are weight and bias parameters for attribute encoding.

  \begin{equation}
  Z_A = \tilde{Z}_A W_A^{(2)} + b_A^{(2)}
  \end{equation}
  where \( Z_A \) is the final latent representation of the node attributes.

- \textbf{Decoder}: Reconstructs the attribute matrix \( \hat{X} \) from \( Z_V \) and \( Z_A \):
  \begin{equation}
  \hat{X} = Z_V (Z_A)^T
  \end{equation}
  where \( \hat{X} \) is the reconstructed attribute matrix.

\paragraph{Loss Function}

The objective is to minimize the reconstruction errors of both structure and attributes:
\begin{equation}
\mathcal{L}_{\text{rec}} = \alpha \| A - \hat{A} \|_F^2 + (1 - \alpha) \| X - \hat{X} \|_F^2
\end{equation}
where \( \alpha \) controls the trade-off between structure \( A \) and attribute \( X \) reconstruction, and \( \| \cdot \|_F^2 \) represents the Frobenius norm.

Nodes with higher reconstruction errors, indicated by the above loss function, are more likely to be anomalies.

\subsubsection{Contrastive Loss Function}

After obtaining the reconstructed adjacency and feature matrices from both the main and augmented graphs, the contrastive loss, which consists of both structural contrastive loss and feature contrastive loss, is applied to these reconstructed matrices. The margin parameter \( m \) used in Equations (12) and (13) is set to 0.5. To further enhance the model’s ability to distinguish between the original and augmented graphs, contrastive learning is incorporated. This allows the network to effectively differentiate between the reconstructions of the main graph and those of the augmented graph containing labeled anomalies.

The contrastive loss is a combination of structural contrastive loss and feature contrastive loss:

\begin{equation}
\mathcal{L}^{sc} = \mathcal{L}_{\text{struct}} + \mathcal{L}_{\text{feat}}
\end{equation}

where:

\begin{equation}
\mathcal{L}_{\text{struct}} = \frac{1}{n} \sum_{i=1}^{n} \left( I_{y_i=0} \cdot d(A_i, \hat{A}_i) + I_{y_i=1} \cdot \max \{0, m - d(A_i, \hat{A}_i) \} \right)
\end{equation}

\begin{equation}
\mathcal{L}_{\text{feat}} = \frac{1}{n} \sum_{i=1}^{n} \left( I_{y_i=0} \cdot d(X_i, \hat{X}_i) + I_{y_i=1} \cdot \max \{0, m - d(X_i, \hat{X}_i) \} \right)
\end{equation}

Where, \( d(\cdot, \cdot) \) represents the distance between matrices, \( I_{y_i=0} \) and \( I_{y_i=1} \) are indicator functions for normal and anomalous instances, and \( m \) is a margin parameter.

The total contrastive loss \( \mathcal{L}^{sc} \) includes:
- \textbf{Positive Pair Loss}: Minimizes the distance between the reconstructed matrices of the main and augmented graphs.
- \textbf{Negative Pair Loss}: Ensures the distance between reconstructed matrices of different graphs exceeds the margin \( m \).

As part of the contrastive learning process, the model evaluates the reconstruction of anomalous nodes in the augmented network during the reconstruction of the main network. The goal is to push the reconstruction of anomalous nodes further from their counterparts in the main network, while pulling the reconstruction of normal nodes closer to their corresponding nodes in the main network. Through these contrasts, the model improves its ability to reconstruct the graph and enhances its anomaly detection capabilities. Additionally, by leveraging the augmented network reconstructions, the model generates a more accurate reconstruction of the main network.

\subsubsection{Total Loss}

The total loss is a weighted sum of the reconstruction loss and the contrastive loss. This combined loss function ensures that the model not only accurately reconstructs the graph but also effectively distinguishes between normal and anomalous graphs.

\begin{equation}
\mathcal{L}_{\text{total}} = \lambda_{\text{rec}} \mathcal{L}_{\text{rec}} + \lambda_{\text{sc}} \mathcal{L}^{sc}
\end{equation}

where \(\lambda_{\text{rec}}\) and \(\lambda_{\text{sc}}\) determine the balance between reconstruction and contrastive losses.

The methodology can be summarized as follows:
1. \textbf{Graph Augmentation}: Create augmented versions of the graph using feature and structural augmentations.
2. \textbf{Reconstruction}: Apply the dual autoencoder to both the main and augmented graphs to reconstruct adjacency and feature matrices.
3. \textbf{Contrastive Learning}: Use contrastive loss to compare the reconstructed matrices, improving anomaly detection by highlighting differences between the main and augmented graphs.

\section{Experiments}

\subsection{Datasets}

We evaluate DCOR using four real-world datasets—Flickr \cite{19}, Amazon \cite{20}, Enron \cite{21}, and Facebook \cite{22}—from diverse domains such as social networks, e-commerce, and email communications.

Table \ref{tab:dataset_statistics} provides key statistics for these datasets.

\begin{table}[ht]
\centering
\caption{Dataset Statistics}
\label{tab:dataset_statistics}
\begin{tabular}{|l|c|c|c|c|}
\hline
\textbf{Dataset} & \textbf{Nodes} & \textbf{Edges} & \textbf{Features} & \textbf{Ground Truth Anomalies} \\
\hline
Flickr & 7,575 & 23,938 & 12,047 & 600 \\
Amazon & 1,418 & 3,695 & 21 & 28 \\
Enron & 13,533 & 176,987 & 18 & 5 \\
Facebook & 4,039 & 88,234 & 576 & 400 \\
\hline
\end{tabular}
\end{table}

\subsection{Experimental Settings}

We evaluate the method on four real-world datasets with consistent hyperparameters across all: 

- Adam optimizer
- Embedding and hidden dimensions set to 128
- Learning rate of 1e-2 (except for Facebook)
- One hidden layer in the encoder

Dataset-specific hyperparameters are detailed in Table \ref{table:hyperparameters}.

\begin{table}[ht]
\centering
\caption{Hyperparameters for each dataset}
\label{table:hyperparameters}
\begin{tabular}{lcccc}
\hline
\textbf{Hyperparameter} & \textbf{Enron} & \textbf{Amazon} & \textbf{Facebook} & \textbf{Flickr} \\
\hline
Epochs & 200 & 200 & 200 & 200 \\
Learning Rate & 0.01 & 0.01 & 0.2 & 0.01 \\
Alpha & 0.1 & 0.5 & 0.1 & 0.9 \\
Structure Anomaly Rate & 0.9 & 0.1 & 0.5 & 0.2 \\
Feature Anomaly Rate & 0.2 & 0.1 & 0.1 & 0.9 \\
Reconstruction  Loss Weight ($\lambda_{\text{rec}}$) & 0.8 & 0.1 & 0.7 & 0.5 \\
Contrastive Loss Weight ($\lambda_{\text{sc}}$) & 0.9 & 0.4 & 0.4 & 0.7 \\
\hline
\end{tabular}
\end{table}

We implemented DCOR using PyTorch and ran the experiments on Google Colab with an NVIDIA T4 GPU and 15GB of memory.
.

\subsection{Experimental Results}

We compare DCOR with several existing methods including LOF \cite{3},  Dominant \cite{6}, AEGIS \cite{23}, AnomalyDAE \cite{5}, and Conad \cite{17}. The AUC scores for anomaly detection are shown in Table \ref{table3}.

\begin{table}[ht]
\centering
\caption{AUC scores of all methods on four datasets.}
\label{table3}
\begin{tabular}{lcccc}
\hline
\textbf{Method} & \textbf{Enron} & \textbf{Amazon} & \textbf{Facebook} & \textbf{Flickr} \\
\hline
LOF & 0.581 & 0.510 & 0.522 & 0.661 \\
Dominant & 0.716 & 0.592 & 0.554 & 0.749 \\
AEGIS & 0.602 & 0.556 & 0.659 & 0.765 \\
AnomalyDAE & 0.552 & 0.611 & 0.741 & 0.694 \\
CONAD & 0.731 & 0.635 & 0.863 & 0.782 \\
\hline
\textbf{DCOR} & \textbf{0.861} & \textbf{0.762} & \textbf{0.926} & \textbf{0.822} \\
\hline
\end{tabular}
\end{table}

To better understand how the various augmentation strategies influence the performance of DCOR, we conducted an ablation study on the Amazon dataset under three different scenarios: feature augmentation only, adjacency augmentation only, and without contrastive learning. We show the results in Table 4, giving the corresponding AUC scores for each augmentation type.

\begin{table}[ht]
\centering
\caption{Ablation study results for DCOR with different augmentation strategies on the Amazon dataset.}
\label{table_ablation}
\begin{tabular}{p{0.5\linewidth} c}
\toprule
\textbf{Augmentation Type} & \textbf{AUC Score} \\
\midrule
Feature Augmentation Only & 0.712 \\
Adjacency Augmentation Only & 0.673 \\
Without Contrastive Learning & 0.592 \\
\bottomrule
\end{tabular}
\end{table}

In this paper, we show that incorporating contrastive learning with reconstruction-based methods provides a more robust anomaly detection approach for attributed networks. Table 3 shows that DCOR consistently outperforms state-of-the-art methods with significantly higher AUC scores on different datasets. The improvement is highly significant in the datasets of Facebook and Enron, where DCOR attained AUC scores of 0.926 and 0.861, respectively. These results indeed promise DCOR as a great potential to enhance anomaly detection, considering its impressively good performance across a variety of settings.

One challenge with traditional contrastive learning methods is that such embeddings may not be able to capture all intricacies within the structure and attributes of the graph, especially when anomalies involve complex patterns. During the process of embedding, subtle details might be overlooked, which in turn reduces anomaly detection accuracy.

Unlike in traditional methods, DCOR does not focus on a comparison of node embeddings but shifts its attention to the reconstruction of adjacency and feature matrices for the graph. This allows the model to learn structural and attribute anomalies that are complex and hard to find otherwise.

Furthermore, DCOR detects effective contrasting views by using customized augmentation techniques; hence, it equips the model for better discrimination between normal and anomaly nodes. This two-dimensional analysis of structure and attributes strengthens the power of anomaly detection, as evidenced by the excellent performance on challenging datasets such as Facebook and Enron.

Finally, the DCOR framework integrates dual reconstruction with contrastive learning to provide an end-to-end anomaly detection approach that significantly enhances its robustness and adaptability to complex network data. This allows the model to accurately reconstruct the graph while effectively distinguishing between normal and abnormal patterns, resulting in superior performance on anomaly detection.

\section{Conclusion}

In this paper, we proposed DCOR a dual contrastive learning based approach for anomaly detection in attributed networks. By contrasting the reconstructed adjacency and feature matrices in both original and augmented graphs, DCOR successfully identified the low-level irregularities likely to be neglected by prior methods. This technique improved the quality of reconstructions in the autoencoder, thereby enabling the model to capture important anomalies.
Extensive experiments conducted on a number of public datasets including Flickr, Amazon, Enron, and Facebook revealed that DCOR outperformed the current state-of-the-art algorithms with higher AUC scores. These results demonstrated the strength of the integrated proposed approach using contrastive learning with reconstruction-based methods for detecting anomalies in attributed networks.

\end{document}